\documentclass{article}
\usepackage{spconf,amsmath,graphicx}
\usepackage{multirow}

\title{SINGLE IMAGE SUPER-RESOLUTION WITH DILATED CONVOLUTION BASED MULTI-SCALE INFORMATION LEARNING INCEPTION MODULE}
\name{Wuzhen Shi, Feng Jiang, Debin Zhao}
\address{School of Computer Science and Technology, Harbin Institute of Technology, Harbin 150001, China}
\begin{document}
%
\maketitle
\begin{abstract}
Traditional works have shown that patches in a natural image tend to redundantly recur many times inside the image, both within the same scale, as well as across different scales. Make full use of these multi-scale information can improve the image restoration performance. However, the current proposed deep learning based restoration methods do not take the multi-scale information into account. In this paper, we propose a dilated convolution based inception module to learn multi-scale information and design a deep network for single image super-resolution. Different dilated convolution learns different scale feature, then the inception module concatenates all these features to fuse multi-scale information. In order to increase the reception field of our network to catch more contextual information, we cascade multiple inception modules to constitute a deep network to conduct single image super-resolution. With the novel dilated convolution based inception module, the proposed end-to-end single image super-resolution network can take advantage of multi-scale information to improve image super-resolution performance. Experimental results show that our proposed method outperforms many state-of-the-art single image super-resolution methods.
\end{abstract}
\section{Introduction}
\label{sec:intro}

\begin{figure}[htb]

\begin{minipage}[b]{1.0\linewidth}
  \centering
  \centerline{\includegraphics[width=8.5cm]{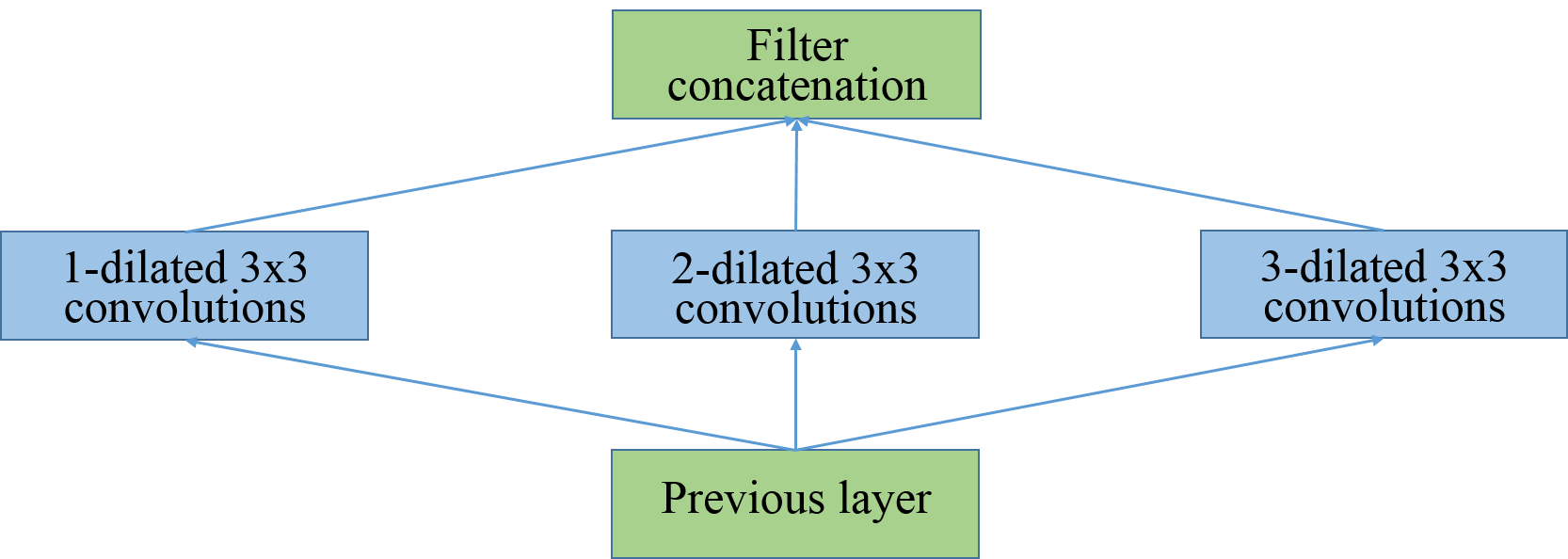}}
  \centerline{(a) Dilated convolution based inception module}\medskip
\end{minipage}
\begin{minipage}[b]{1.0\linewidth}
  \centering
  \centerline{\includegraphics[width=8.5cm]{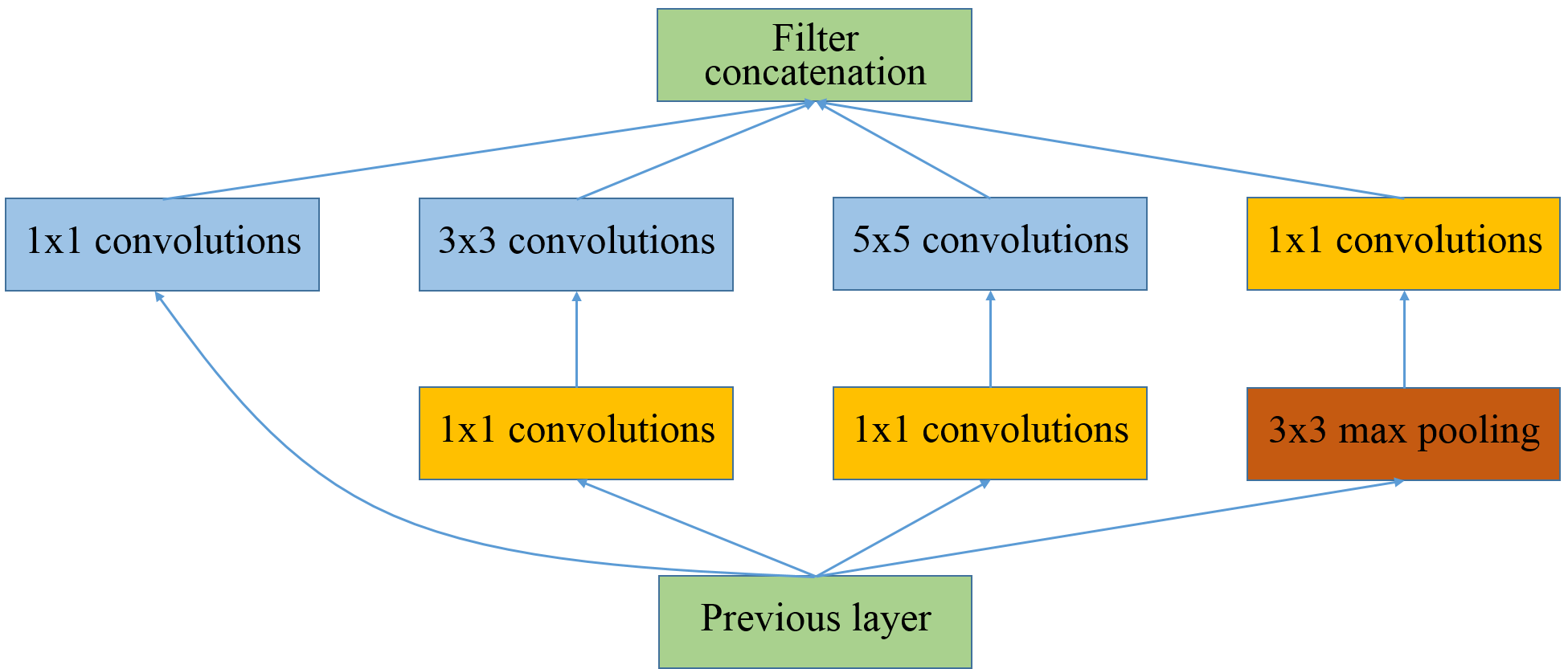}}
  \centerline{(b) Original GoogLeNet inception module~\cite{ref14}}\medskip
\end{minipage}
\vspace{-0.98cm}
\caption{Comparison between the proposed dilated convolution based inception module and the original GoogLeNet inception module. Our proposed new inception module contains multiple different scale dilated convolution that makes it can learn multi-scale image information.}
\label{fig:comparison}
\vspace{-0.6cm}
\end{figure}

In this paper, we focus on single image super-resolution (SR), which aims at recovering a high-resolution (HR) image from a given low-resolution (LR) one. It is always the research emphasis because of the requirement of higher definition video displaying, such as the new generation of Ultra High Definition (UHD) TVs. However, most video content would not at sufficiently high resolution for UHD TVs. As a result, we have to develop efficient SR algorithms to generate UHD content from lower resolutions \cite{ref1}.

Traditional single image SR methods always try to find a new image prior or propose a new way to use the existing image priors. A lot of image prior information have been explored in the image restoration literature, such as local smooth, non-local self-similarity and image sparsity. Based on the assumption that low and high resolution images have the same sparse representation, Yang et al. \cite{ref2} use two coupled dictionaries to learn a nonlinear mapping between the LR and the HR images. In \cite{ref3}, Glasner et al. begin to use the image multi-scale information for single image SR and obtain state-of-the-art results.

Recently, due to the availability of much larger training dataset and the powerful GPU implementation, deep learning based methods achieve great success in many fields, including both high level and low level computer vision problems. Look through the literature, most state-of-the-art single image SR methods are based on deep learning. The pioneering SR method is SRCNN proposed by Dong et al. \cite{ref4, ref5}. They establish the relationship between the traditional sparse-coding based SR methods and their network structure and demonstrate that a convolutional neural network (CNN) can learn a mapping from low resolution image to high resolution one in an end-to-end manner. Dong et al. successfully expand SRCNN for compression artifacts reduction by introducing a feature enhancement layer \cite{ref6}. Soon after, they proposed a fast SRCNN method, which directly maps the original low-resolution image to the high-resolution one \cite{ref7}. Different from \cite{ref4,ref5,ref6,ref7}, some works try to learn image high frequency details instead of the undegraded image. In \cite{ref9}, Kim et al. cascade many convolutional layers to form a very deep network to learn image residual.

Investigating an effective way to use multi-scale information is also important. The degraded image can be successful recovered is mainly based on the assumption that patches in a natural image tend to redundantly recur many times inside the image. However, it is not only exist in the same scale but also across different scales. It has been demonstrated that make full use of multi-scale information can improve the restoration result in traditional methods \cite{ref3}. However, the multi-scale information has been little investigated in deep learning methods. In \cite{ref9}, Kim et al. try to train a multi-scale model for different magnification SR. It is a very rough tactics to exploit the scale information since they just put different scale image as input for training. Its success can be attribute to the powerful learning ability of CNN instead of the multi-scale information being considered in the network structure.

In this paper, we propose a dilated convolution based inception module to learn multi-scale information and design a deep network for single image SR. Fig. \ref{fig:comparison} makes a comparison between the proposed dilated convolution based inception module and the original GoogLeNet inception module. Our proposed new inception module contains multiple different scale dilated convolution that makes it can learn multi-scale image information. Furthermore, we cascade multiple dilated convolution based inception modules to constitute a deep network for single image SR. In short, the contributions of this work are mainly in three aspects: 1) we proposed a dilated convolution based inception module, which can learn multi-scale information with only single scale image input; 2) we design a novel deep network with the proposed dilated convolution based inception module for single image SR. 3) experimental results show that our proposed new method outperforms many state-of-the-art methods.

\section{DILATED CONVOLUTION BASED INCEPTION MODULE}
\label{sec:dilated}

\begin{figure}[htb]

\begin{minipage}[b]{1.0\linewidth}
  \centering
  \centerline{\includegraphics[width=7.5cm]{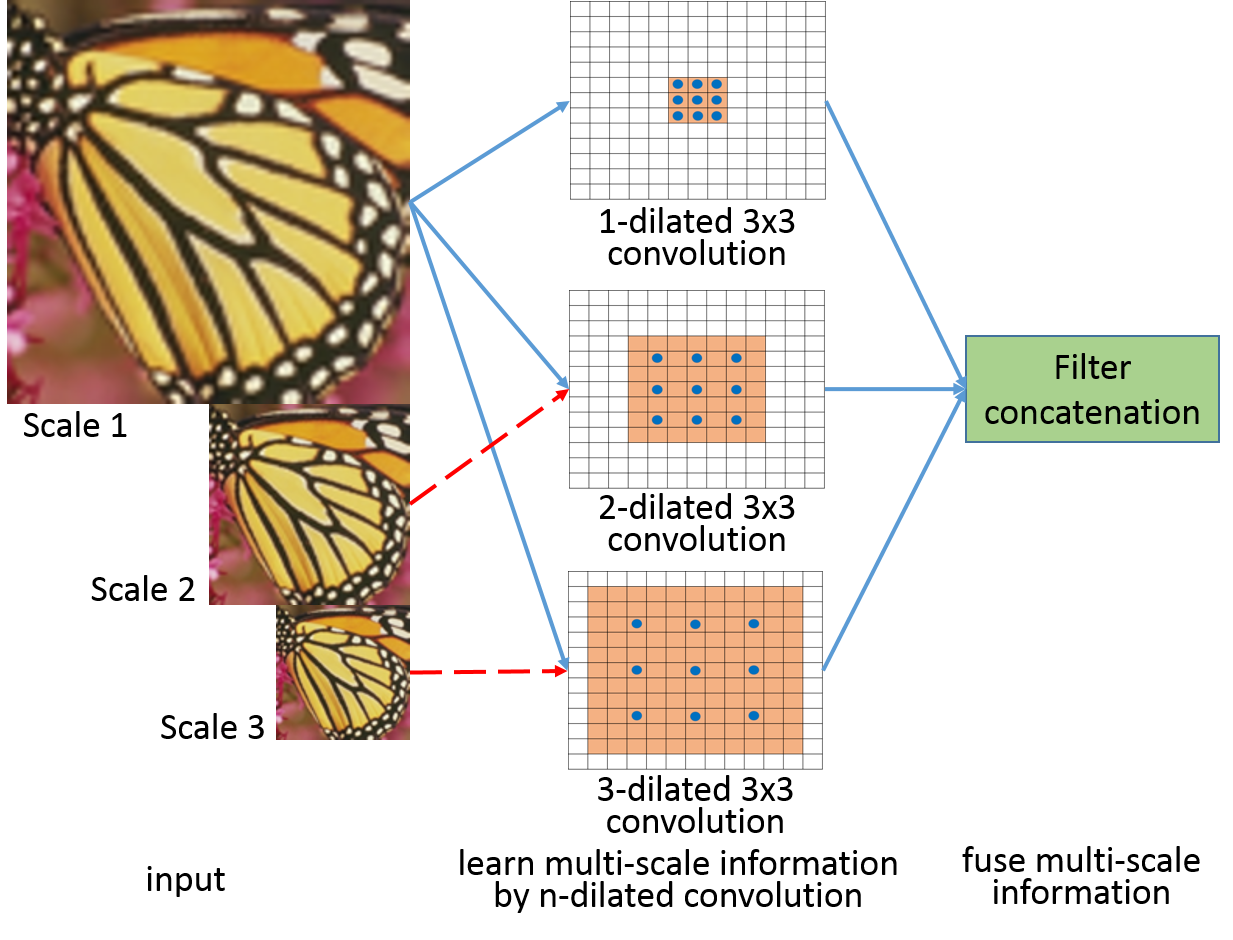}}
\end{minipage}
\vspace{-0.95cm}
\caption{Illustration of the proposed inception how to learn multi-scale information.}
\label{fig:illustration}
\vspace{-0.5cm}
\end{figure}

\textbf{Dilated Convolution}: It has been referred to in the past as “convolution with a dilated filter” \cite{ref11, ref12}. In \cite{ref13}, Yu et al. call it “dilated convolution” instead of “convolution with a dilated filter” to clarify that no “dilated filter” is constructed or represented. It can be formulated as
\begin{eqnarray}
\left( {F{ * _l}k} \right)\left( p \right) = \sum\nolimits_{s + lt = p} {F\left( s \right)k\left( t \right)}
\end{eqnarray}
where ${{ * _l}}$ is called as a dilated convolution or an $l$-dilated convolution, $F$ and $k$ is a discrete function and a discrete filter of size ${\left( {2r + 1} \right)^2}$, respectively. Three dilated convolutions have been shown in the middle part of Fig. \ref{fig:illustration}, where $k$ is a $3 \times 3$ filter and the kernel dilation factors are 1, 2 and 3, respectively. It shows that filters are dilated by inserting $dilated - 1$ zeros between filter elements for a given dilation factor. We refer the reader to \cite{ref13} for more information about the dilated convolution.

\textbf{Dilated Convolution based Inception Module}: To make full use of the image multi-scale information, Szegedy et al. \cite{ref14} proposed an inception module for image classification. As shown in Fig. \ref{fig:comparison}b, the GoogLeNet inception module contains multiple convolution with different kernel size. It concatenates the outputs of these different size filter to fuse different scale information. With this multi-scale information learning structure, GoogLeNet achieves the new state-of-the-art performance for classification and detection in the ImageNet Large-Scale Visual Recognition Challenge 2014 (ILSVRC14). Inspired by this successful work, we propose a dilated convolution based inception module to learn multi-scale information for improving single image SR performance.

As shown in Fig. \ref{fig:comparison}a, our new inception module first use three dilated convolution with kernel size $3 \times 3$ and dilated factor 1, 2 and 3, respectively, to operate on the previous layers. Then, we concatenate all these convolution outputs to fuse different scale information.

\textbf{Insights}: there are a lot of ways to learn the multi-scale information by deep network. For example, the GoogLeNet inception module uses different kernel size of convolution with different receptive field to operate on the previous layer. However, it is still operated on the same image scale space. The other choice is to operate on different scale image input. Farabet et al. \cite{ref15} use this way to learn multi-scale feature for scene parsing. For single image SR, the lower solution image always has much more sharp detail information than the interpolation result. Therefore, it can improve the SR result by taking these small scale images into account. Fig. \ref{fig:illustration} shows the process of our proposed dilated convolution based inception module how to learn multi-scale information. The blue solid lines indicate that all these dilated convolutions actually operate on the same scale image, while the red dash lines mean these convolutions with different dilated factors learn the corresponding scale image information. Furthermore, the output of the dilated convolution can keep the same size with its input, so we can fuse these different scale information through concatenation operator easily.

\section{PROPOSED MULTI-SCALE INFORMATION LEARNING NETWORK STRUCTURE}
\label{sec:proposed}

\begin{figure*}[htb]

\begin{minipage}[b]{1.0\linewidth}
  \centering
  \centerline{\includegraphics[width=0.95\textwidth]{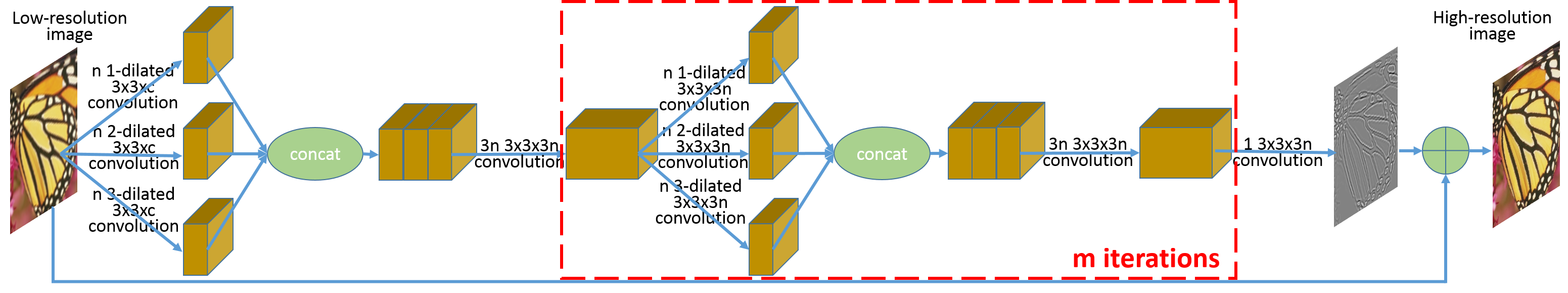}}
\end{minipage}
\vspace{-0.9cm}
\caption{The proposed single image super-resolution network structure with dilated convolution based multi-scale information learning inception module. It uses inception module, which contains different scale dilated convolution, to learn multi-scale information. Multiple inception modules cascade to constitute a deep network to predict high frequency detail information.}
\label{fig:networkframework}
\vspace{-0.4cm}
\end{figure*}

The configuration of our proposed single image SR network structure is outlined in Fig. \ref{fig:networkframework}, which cascades $m + 1$ dilated convolution based inception modules and $m + 2$ common discrete convolutions. It can be explained as three phases, i.e. feature extraction, feature enhancement and image reconstruction. Since residual learning have been proven to be an effective way to accelerate the network convergence in recent works \cite{ref9}, we follow them to make our network predict image high frequency details instead of the HR image itself. The predicted image frequency details will be added to the input LR image to get the final desired HR one.

As most single image SR works done, we first up-sample the LR image to the desired size using bicubic interpolation. For the ease of presentation, we still call the interpolation result as a "low-resolution" image, although it has the same size as the HR image. For the feature extraction phase, $n$ dilated convolution based inception modules operate on the LR input. The filter kernel size is $3 \times 3 \times c$, where $c$ is the number of image channel, for the first inception module layer. The inception module combines different scale feature information through concatenation operator. To further fuse these multi-scale information, in the next we add a nonlinear transformation layer after the inception module layer, which contains $3n$ common convolution with filter kernel size $3 \times 3 \times 3n$. 

Both the high level and low level vision works have proven the deeper the network the better the performance. Furthermore, the larger reception field of the network can catch more contextual information that gives more clues for predicting high frequency details. Therefore, we iterate $m$ times the process of dilated convolution based inception modules following with a common convolution operator for the feature enhancement phase. In this phase, the filter kernel size is $3 \times 3 \times 3n$ as show in Fig. \ref{fig:networkframework}. In the image reconstruction phase, we use a single common convolution of size $3 \times 3 \times 3n$ to predict the high frequency details. Finally, the predicted high frequency details will add to the interpolation output to get the desired HR image.

\subsection{Training}
\label{subsec:training}

There are a lot of perceptually relevant characteristics based loss functions have been proposed in the literature. But for a fair comparison with SRCNN and FSRCNN, we adopt the mean square error (MSE) as the cost function of our network. Our goal is to train an end-to-end mapping $F$ to predict high frequency detail $\hat y = F\left( x \right)$, where $x$ is an interpolation result of the LR image and $\hat y$ is the estimated high frequency detail image. Given a training dataset $\left\{ {{x_i},{y_i}} \right\}_{i = 1}^N$, the optimization objective is represented as
\begin{eqnarray}
\mathop {\min }\limits_\theta  \frac{1}{{2N}}\sum\nolimits_{i = 1}^N {\left\| {F\left( {{x_i};\theta } \right) - {y_i}} \right\|} _F^2
\end{eqnarray}
where $\theta $ is the network parameters needed to be trained, ${F\left( {{x_i};\theta } \right)}$ is the estimated high frequency image with respect to the interpolation result of a LR image. In the literature, people suggest to use to recently proposed Parametric Rectified Linear Unit (PReLU) as the activation function instead of the commonly-used Rectified Linear Unit (ReLU). But, in order to reduce parameters, ReLU is used after each convolution layer in our very deep network that has got a satisfactory result for comparison. We use the adaptive moment estimation (Adam) \cite{ref16} to optimize all network parameters instead of the commonly used stochastic gradient descent one.

\section{EXPERIMENTAL RESULTS}
\label{sec:experimental}

\begin{table*}[]
\small
\centering
\caption{The results of average PSNR (dB) and SSIM on the Set5 \cite{ref18}, Set14 \cite{ref19} and BSD200 \cite{ref20} dataset}
\label{tab:quantitative}
\begin{tabular}{|c|c|c|c|c|c|c|c|c|}
\hline
\multirow{2}{*}{Dataset} & \multirow{2}{*}{Scale} & Bicubic      & A+           & SRF          & SRCNN        & SCN          & FSRCNN       & MSSRNet               \\ \cline{3-9} 
                         &                        & PSNR/SSIM    & PSNR/SSIM    & PSNR/SSIM    & PSNR/SSIM    & PSNR/SSIM    & PSNR/SSIM    & PSNR/SSIM             \\ \hline
\multirow{3}{*}{Set5}    & $ \times 2$                      & 33.66/0.9299 & 36.55/0.9544 & 36.87/0.9556 & 36.34/0.9521 & 36.76/0.9545 & 37.00/0.9558 & \textbf{37.33/0.9581} \\ \cline{2-9} 
                         & $ \times 3$                     & 30.39/0.9299 & 32.59/0.9088 & 32.71/0.9098 & 32.39/0.9033 & 33.04/0.9136 & 33.16/0.9140 & \textbf{33.38/0.9178} \\ \cline{2-9} 
                         & $ \times 4$                      & 28.42/0.8104 & 30.28/0.8603 & 30.35/0.8600 & 30.09/0.8503 & 30.82/0.8728 & 30.71/0.8657 & \textbf{31.10/0.8777} \\ \hline
\multirow{3}{*}{Set14}   & $ \times 2$                      & 30.23/0.8687 & 32.28/0.9056 & 32.51/0.9074 & 32.18/0.9039 & 32.48/0.9067 & 32.63/0.9088 & \textbf{32.89/0.9117} \\ \cline{2-9} 
                         & $ \times 3$                      & 27.54/0.7736 & 29.13/0.8188 & 29.23/0.8206 & 29.00/0.8145 & 29.37/0.8226 & 29.43/0.8242 & \textbf{29.57/0.8282} \\ \cline{2-9} 
                         & $ \times 4$                      & 26.00/0.7019 & 27.32/0.7471 & 27.41/0.7497 & 27.20/0.7413 & 27.62/0.7571 & 27.59/0.7535 & \textbf{27.83/0.7631} \\ \hline
\multirow{3}{*}{BSD200}  & $ \times 2$                      & 29.70/0.8625 & 31.44/0.9031 & 31.65/0.9053 & 31.38/0.9287 & 31.63/0.9048 & 31.80/0.9074 & \textbf{32.08/0.9118} \\ \cline{2-9} 
                         & $ \times 3$                      & 27.26/0.7638 & 28.36/0.8078 & 28.45/0.8095 & 28.28/0.8038 & 28.54/0.8119 & 28.60/0.8137 & \textbf{28.78/0.8188} \\ \cline{2-9} 
                         & $ \times 4$                      & 25.97/0.6949 & 26.83/0.7359 & 26.89/0.7368 & 26.73/0.7291 & 27.02/0.7434 & 26.98/0.7398 & \textbf{27.17/0.7489} \\ \hline
\multicolumn{2}{|c|}{Avg.}                        & 28.80/0.8151 & 30.53/0.8491 & 30.67/0.8505 & 30.39/0.8474 & 30.81/0.8542 & 30.88/0.8537 & \textbf{31.13/0.8596} \\ \hline
\end{tabular}
\vspace{-0.5cm}
\end{table*}

A lot of experiments have been done to show dramatic improvement in performance can be obtained by our proposed method. We give the experimental details and report the quantitative results on three popular dataset in this section. We name the proposed method as Multiple Scale Super-Resolution Network (MSSRNet).

\subsection{Datasets for Training and Testing}
\label{subsec:datasets}

It is well known that training dataset is very important for the performance of learning based image restoration methods. A lot of training dataset can be found in the literature. For example, SRCNN \cite{ref4,ref5} uses a 91 images dataset and VDSR \cite{ref9} uses 291 images dataset. However, the 91 images dataset is too small to push a deep model to the best performance. Some image in the 291 image dataset are JPEG format, which are not optimal for the SR task. We follow FSRCNN \cite{ref7} to use the General-100 dataset, which contains 100 bmp-format images (with no compression). We set the patch size as $48 \times 48$, and use data augmentation (rotation or flip) to prepare training data. Following FSRCNN, SRCNN and SCN, we use three dataset, i.e. Set5 \cite{ref18} (5 images), Set14 \cite{ref19} (14 images) and BSD200 \cite{ref20} (200 images) for testing, which are widely used for benchmark in other works.

\subsection{Implementing Details and Parameters}
\label{subsec:implementing}

For each dilated convolution based inception module layer, we set $n = 8$. That is, there are 8 inception module per layer. For feature enhancement process, we iterate 5 inception modules to make the network deeper, i.e. $m = 5$. Filters are initialized using the initializer proposed by He et al. \cite{ref21} with values sampled from the Uniform distribution. For the other hyper-parameters of Adam, we follow \cite{refadd1} to set the exponential decay rates for the first and second moment estimate to 0.9 and 0.999, respectively. Each model was trained only 100 epochs and each epoch iterates 2000 times with batch size of 64. We set a lager learning rate in the initial training phase to accelerate convergence, then decrease it gradually to make the model more stable. Therefore, the learning rates are 0.001, 0.0001 and 0.00001 for the first 50 epochs, the 51 to 80 epochs and the last 20 epochs, respectively. Our model is implemented by the MatConvNet package [17].

\subsection{Comparisons with State-of-the-Art Methods}
\label{subsec:comparisons}

We compare our method with five state-of-the-art learning based SR algorithms that rely on external databases, namely the A+ \cite{ref22}, SRF \cite{ref23}, SRCNN \cite{ref4,ref5}, FSRCNN \cite{ref7} and SCN \cite{ref24}. A+ and SRF are two state-of-the-art traditional methods, while SRCNN, FSRCNN and SCN are three newest popular deep learning based single image super-resolution image methods. In Table \ref{tab:quantitative}, we provide a summary of quantitative evaluation on several datasets. The results of other five methods are the same as reported at FSRCNN. Our method outperforms all previous methods in these datasets. Compare with the newest FSRCNN, our method can improve roughly 0.33 dB, 0.22Db and 0.37 dB on average with respect to up-sample factor 2, 3 and 4 on Set5 dataset, respectively. Over the three dataset and three up-sample factor, our MSSRNet can improve roughly 2.33 dB, 0.6 dB, 0.46 dB, 0.74 dB, 0.32 dB and 0.25 dB on average, in comparison with Bicubic, A+, SRF, SRCNN, SCN and FSRCNN, respectively. To get better performance, we can increase the network depth (larger $m$), which is called deeper is better in the literature, and the network width with larger $n$. In our experiments, we have implemented the fatter network with $n = 16$ and $n = 32$, and the deeper network with $m = 10$ and $m = 15$. Both the deeper and the fatter networks show PSNR and SSIM gain. The reader can download our test code\footnote{https://github.com/wzhshi/MSSRNet}  to get more quantitative and qualitative results.

\section{CONCLUSION}
\label{sec:conclution}

In this paper, we use deep learning technology to solve the single image super-resolution problem. We first propose a dilated convolution based inception module, which can learn multi-scale information from the single scale input image. We design a deep network, which cascades multiple dilated convolution based inception modules, for single image super-resolution. Experimental results show that the proposed method outperforms many state-of-the-art ones. As future work we plan to explore MSSRNet for video processing.

\section{ACKNOWLEDGEMENTS}
\label{sec:acknowledgements}

This work has been supported in part by the Major State Basic Research Development Program of China (973 Program 2015CB351804), the National Science Foundation of China under Grant No. 61572155.

\bibliographystyle{IEEEbib}

\bibliography{refs}

\end{document}